\documentclass[]{fairmeta}

\usepackage{booktabs}
\usepackage{multicol}
\usepackage{multirow}
\usepackage{color}
\usepackage{colortbl}
\usepackage{caption}
\usepackage{hhline}
\usepackage{pifont}
\usepackage{threeparttable}
\usepackage{makecell}
\usepackage{listings}
\usepackage{amsfonts,amssymb}
\usepackage{graphicx}
\usepackage[export]{adjustbox}
\usepackage{lipsum}

\newcommand\blfootnote[1]{%
  \begingroup
  \renewcommand\thefootnote{}\footnote{#1}%
  \addtocounter{footnote}{-1}%
  \endgroup
}
\usepackage{enumitem}

\usepackage{color, colortbl}
\definecolor{citecolor}{HTML}{0071bc}

\newcommand\hoo{\raisebox{-0.2em}{\includegraphics[width=1.2em]{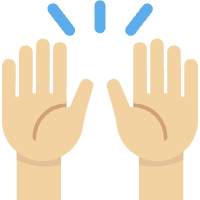}}}
\newcommand\aria{\raisebox{-0.3em}{\includegraphics[width=1.5em]{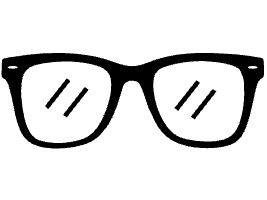}}}
\newcommand\assembly{\raisebox{-0.3em}{\includegraphics[width=1.3em]{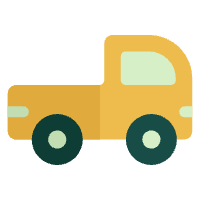}}}
\newcommand\object{\raisebox{-0.3em}{\includegraphics[width=1.5em]{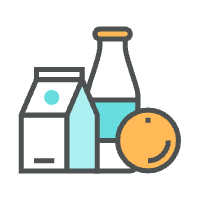}}}
\newcommand\subject{\raisebox{-0.4em}{\includegraphics[width=1.3em]{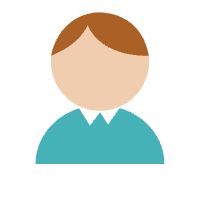}}}
\newcommand\scene{\raisebox{-0.3em}{\includegraphics[width=1.2em]{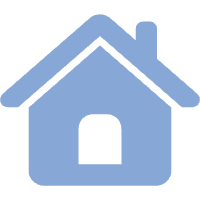}}}

\usepackage{colortbl}
\newcommand{\best}[1]{\textcolor{black}{\textbf{#1}}}

\newcommand{\thickhline}{\Xhline{2\arrayrulewidth}}
\newif\ifclean

\definecolor{Gray}{gray}{0.9}
\definecolor{highlight}{gray}{0.9}
\newcommand{\custompar}[1]{
  \par
  \vspace{2pt}
  \noindent\textbf{#1}
}

\title{Put Myself in Your Shoes: Lifting the Egocentric Perspective from Exocentric Videos}

\author[1]{Mi Luo}
\author[1,2]{Zihui Xue}
\author[1]{Alex Dimakis}
\author[1,2]{Kristen Grauman}

\affiliation[1]{The University of Texas at Austin}
\affiliation[2]{FAIR at Meta}

\abstract{We investigate exocentric-to-egocentric cross-view translation, which aims to generate a first-person (egocentric) view of an actor based on a video recording that captures the actor from a third-person (exocentric) perspective. To this end, we propose a generative framework called Exo2Ego that decouples the translation process into two stages: high-level structure transformation, which explicitly encourages cross-view correspondence between exocentric and egocentric views, and a diffusion-based pixel-level hallucination, which incorporates a hand layout prior to enhance the fidelity of the generated egocentric view. To pave the way for future advancements in this field, we curate a comprehensive exo-to-ego cross-view translation benchmark. It consists of a diverse collection of synchronized ego-exo tabletop activity video pairs sourced from three public datasets: H2O, Aria Pilot, and Assembly101. The experimental results validate that Exo2Ego delivers photorealistic video results with clear hand manipulation details and outperforms several baselines in terms of both synthesis quality and generalization ability to new actions.}

\date{\today}

\usepackage{hyperref}
\usepackage{amsmath}
\usepackage{comment}
\usepackage{multirow,bigdelim}
\usepackage{lipsum}
\usepackage{array}
\usepackage{wrapfig}

\usepackage{adjustbox}
\usepackage{overpic}
\usepackage{makecell}
\usepackage[normalem]{ulem}
\usepackage{blindtext}
\usepackage{xcolor}
\usepackage{soul}
\usepackage{amsfonts}
\usepackage[linesnumbered, boxed, ruled]{algorithm2e}
\newcolumntype{C}[1]{>{\centering\let\newline\\\arraybackslash\hspace{0pt}}m{#1}}

\newif\ifdraft
\drafttrue

\ifdraft
\definecolor{darkg}{rgb}{0,0.4,0}
\newcommand{\nrc}[1]{{\color{red}[\textbf{Nataniel:} #1]}}
\newcommand{\mrc}[1]{{\color{purple}[\textbf{Miki:} #1]}}
\newcommand{\vjc}[1]{{\color{blue}[\textbf{Varun:} #1]}}
\newcommand{\kac}[1]{{\color{teal}[\textbf{Kfir:} #1]}}
\newcommand{\yzc}[1]{{\color{violet}[\textbf{Yuanzhen:} #1]}}
\newcommand{\ypc}[1]{{\color{darkg}[\textbf{Yael:} #1]}}

\newcommand{\drop}[1]{}

\else
\newcommand{\nrc}[1]{}
\newcommand{\mrc}[1]{}
\newcommand{\vjc}[1]{}
\newcommand{\kac}[1]{}
\newcommand{\yzc}[1]{}
\newcommand{\ypc}[1]{}

\fi

\makeatletter
\DeclareRobustCommand\onedot{\futurelet\@let@token\@onedot}
\def\@onedot{\ifx\@let@token.\else.\null\fi\xspace}

\makeatother

\usepackage[bottom]{footmisc}
\usepackage{arydshln}

\usepackage{enumitem}

\makeatletter
\def\blfootnote{\xdef\@thefnmark{}\@footnotetext}
\makeatother

\newif\ifwatermark
\watermarktrue
\draftfalse

\usepackage{amsmath,amsfonts,bm}
\usepackage{mathtools}

\def\eqref#1{equation~\ref{#1}}

\def\1{\bm{1}}

\DeclareMathAlphabet{\mathsfit}{\encodingdefault}{\sfdefault}{m}{sl}
\SetMathAlphabet{\mathsfit}{bold}{\encodingdefault}{\sfdefault}{bx}{n}

\newcommand{\E}{\mathbb{E}}

\newcommand{\Eb}[2]{\E_{#1}\!\left[#2\right]}

\newcommand{\bI}{\mathbf{I}}

\newcommand{\bzero}{\mathbf{0}}

\newcommand{\bd}{\mathbf{d}}

\newcommand{\bz}{\mathbf{z}}

\newcommand{\bepsilon}{{\boldsymbol{\epsilon}}}

\begin{document}

\maketitle
\section{Introduction}
\begin{figure}[t]
\centering
\includegraphics[scale=1.1]{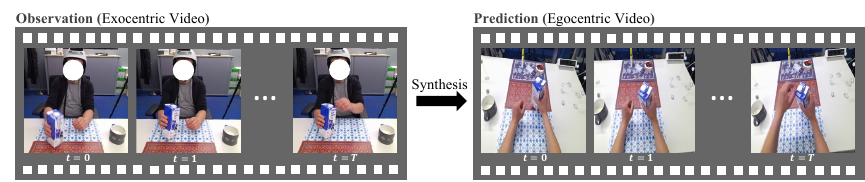}
\caption{
Cross-view translation from exocentric to egocentric video. Given an exocentric video sequence (top), the goal is to generate the corresponding egocentric perspective (bottom).}
\label{fig:setting}
\end{figure}

Given a third-person video capturing a person opening a milk carton, what would the visual world look like from his perspective?  See Figure~\ref{fig:setting}. 
Due to mirror neurons in the human brain~\cite{ardeshir2018exocentric}, we can easily envision the appearance and spatial relationships of the person's hands and the milk carton from a first-person perspective. 
However, existing computer vision models struggle to do the same thing, owing to the stark distinctions between the two viewpoints. 

We make a step towards addressing this underexplored problem: exocentric-to-egocentric cross-view translation. The goal is to synthesize the corresponding ego view of an actor from an exo video recording\footnote{We use ``ego'' and ``exo'' as shorthand for egocentric (first-person) and exocentric (third-person).}, with minimal assumptions on the viewpoint relationships (e.g., camera parameters or accurate geometric scene structure). Specifically, we focus on synthesizing ego tabletop activities that involve significant hand-object interactions, such as assembling toys or pouring milk.

This task can benefit many applications, ranging from robotics to virtual and augmented reality (VR/AR). 
For example, an AR assistant could transform the view shown in a third-person how-to video to demonstrate to a user how things should look from their own perspectives, e.g., revealing finger placements and strumming techniques for a guitar video.
Similarly, in robot learning, the robot could better match its actions to those of a human instructor in the room by projecting his/her hand-object interactions to its own perspective.  Indeed, recent work demonstrates that the human ego view is valuable for robot learning~\cite{bharadhwaj2023zero, bahl2022human, majumdar2023we, nair2022r3m, mandikal2021dexvip}. 

However, exo-to-ego view translation is highly challenging.  It requires understanding the spatial relationships of visible hands and objects and inferring their pixel-level appearance in the novel ego view. Further, the task is not strictly geometric; it is inherently underdetermined.  
Some parts of an object may not be visible in the exo view---such as the inner pages of a book when only the cover is observed in the exo view---requiring the model to extrapolate the occluded parts. As a result, the recent popular geometry-aware novel view synthesis approaches~\cite{mildenhall2020nerf, niemeyer2022regnerf, Yu_2021_CVPR, jang2021codenerf} are ill-equipped to solve this problem. The key reason is that they are regressive rather than generative, which limits their ability to deal with sparse input views (single exo view in our case) and major occlusions.

In light of this, we propose a probabilistic exo-to-ego generative framework (termed Exo2Ego) that learns the conditional distribution of the target ego video given the exo video. Exo2Ego differs from generative-based methods for general novel view synthesis~\cite{liu2021infinite, ren2022look, rombach2021geometry, wiles2020synsin, watson2022novel, tseng2023consistent, chan2023generative} as it relaxes the requirement for camera parameters as input.
This expands the potential application scenarios since camera parameters are rarely accessible in real-world scenarios, e.g., taking photos or videos with a mobile phone. Additionally, accurately inferring camera parameters remains challenging without precise correspondences or a rigid lab setting for calibration~\cite{wang2021nerfmm}.
Moreover, Exo2Ego stands apart from recent cross-view image-to-image translation methods~\cite{tang2019selectiongan, ren2021crossmlp}, as it does not require the ground truth semantic map (object segmentation) of the target domain at the inference stage---typically an infeasible assumption. 

Our key insights are to explicitly encourage cross-view correspondence by predicting the ego layout and to introduce a hand layout prior to boost the fidelity of generated hands in the ego view.
Specifically, our Exo2Ego framework decouples the exo-to-ego view translation problem into two stages: (1) High-level Structure Transformation, which infers the rough location and interaction manner of the hands and objects in the ego view, using a transformer-based model to transform the exo hand-object interaction layout into its ego counterpart; and (2) Diffusion-based Pixel Hallucination, which trains a conditional diffusion model to refine the details on top of the ego hand-object interaction layout.
Overall, Exo2Ego is a generative framework that offers a simple but effective baseline approach for the exo-to-ego view translation problem, accounting for the centrality of hand-object manipulations in this domain.

To evaluate our approach, we construct an exo-to-ego cross-view synthesis benchmark, comprising synchronized ego-exo video pairs curated from three datasets: H2O~\cite{Kwon_2021_ICCV}, Aria Pilot~\cite{aria_pilot_dataset}, and Assembly101~\cite{sener2022assembly101}. Empirical results underscore the efficacy of our Exo2Ego framework, which produces realistic video outputs with distinct hand manipulation details. 
Exo2Ego surpasses single-view translation baselines~\cite{wang2018high, wang2018video}, a recent cross-view synthesis approach~\cite{liu2020pgan}, as well as a NeRF-based method~\cite{Yu_2021_CVPR}, demonstrating superior generation quality and a marked improvement in generalization capability to new actions.

\section{Related work}
\begin{figure*}[h]
\centering
\includegraphics[scale=0.58]{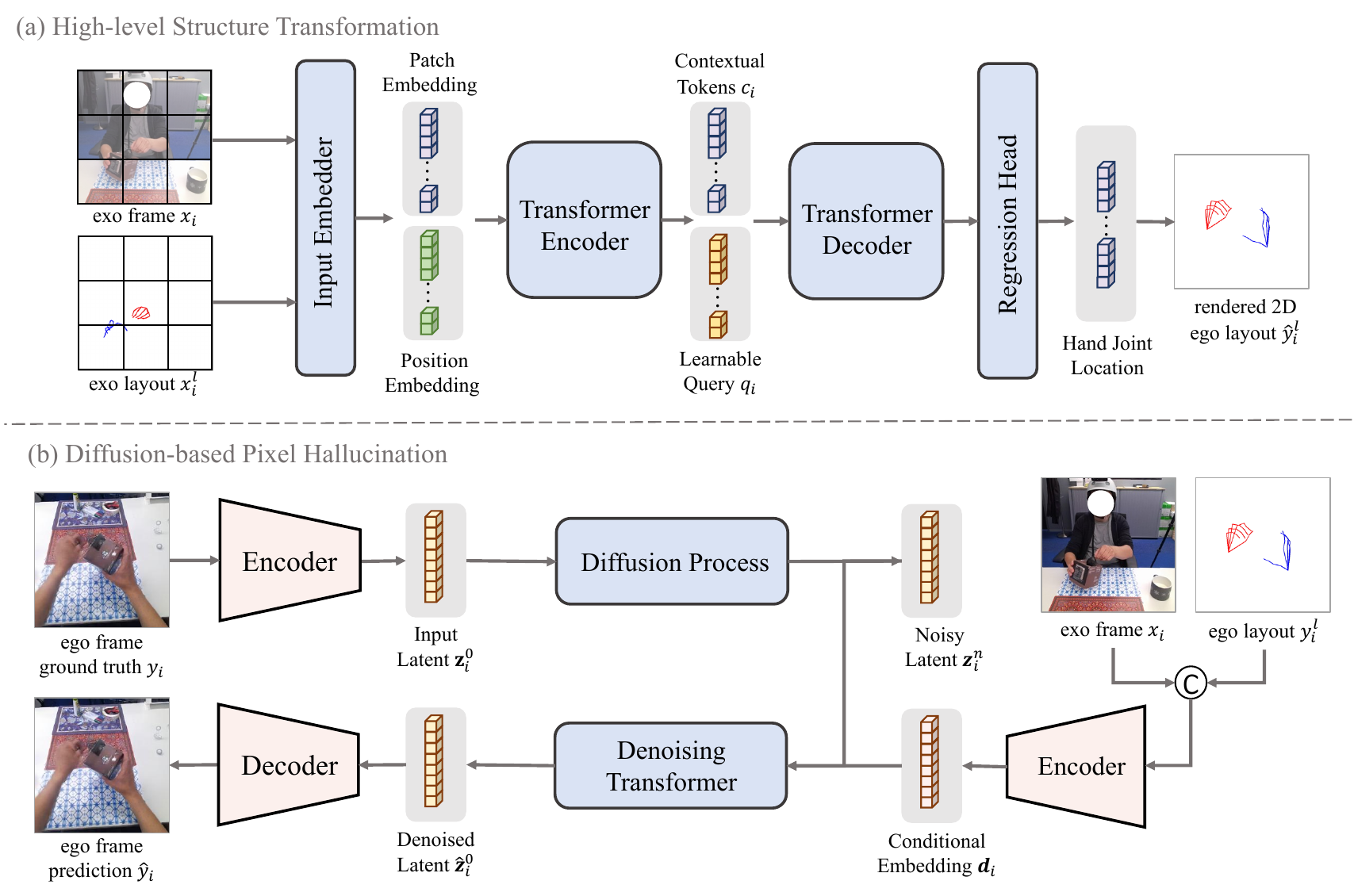}
\caption{Our Exo2Ego framework comprises two modules: (a) High-level Structure Transformation, which predicts the ego layout, capturing hand position and interactions using a transformer-based encoder-decoder architecture. (b) Diffusion-based Pixel Hallucination, which enhances pixel-level details on top of the ego layout using a conditional diffusion model. 
}
\label{fig:framework}
\end{figure*}

\paragraph{Relating Egocentric and Exocentric Views}
There have been several attempts to jointly understand ego and exo views~\cite{elfeki2018third, sigurdsson2018charades, grauman2023ego}. 
Early efforts explore how to localize the person wearing a camera in a third-person view, given their egocentric video~\cite{ardeshir2016ego2top, ardeshir2018egocentric, fan2017identifying, xu2018joint, Wen_2021_ICCV}.  To bridge the ego-exo gap, other work learns view-invariant features from concurrent (paired)~\cite{ardeshir2018exocentric, sigurdsson2018actor, Sermanet2017TCN, yu2019see, yu2020first} or unpaired~\cite{sherry-neurips2023} views, or boosts the latent ego signals in exo video during pretraining~\cite{li2021ego}, summarization~\cite{ho2018summarizing}, and 3D pose estimation~\cite{wang2022estimating}.
Additionally, fusing ego and exo views can improve action recognition~\cite{soran2015action} and robotic manipulation tasks~\cite{jangir2022look}.

Despite its significance, cross-view translation has received little attention. 
P-GAN~\cite{liu2020pgan} and STA-GAN~\cite{liu2021stagan} have explored cross-view image and video synthesis, respectively, yet 
they only examine basic activities such as walking, jogging, and running, limiting their applicability to more diverse scenarios. 
To address these limitations, we introduce a new benchmark for exo-to-ego view synthesis covering some diverse activities, from assembling toys to manipulating everyday objects. Importantly, we relax STA-GAN's requirement for the ego semantic map during test time~\cite{liu2021stagan}, enhancing the practical applicability.

\vspace*{-0.12in}
\paragraph{Novel View Synthesis}
Exo-to-ego view translation also relates to novel view synthesis, i.e., given a set of images of a scene, infer how the scene looks from novel viewpoints.
State-of-the-art geometry-aware approaches like Scene Representation Networks (SRN)~\cite{sitzmann2019scene} and Neural Radiance Fields (NeRF)~\cite{mildenhall2020nerf, niemeyer2022regnerf, Yu_2021_CVPR, jang2021codenerf} learn 3D representations from 2D images and camera parameters and perform differentiable neural rendering from one or a few input images at inference. Although they excel at interpolating near-input views, they are typically not well-suited for the large view changes required for exo-to-ego view synthesis. 
Trained purely with regression objectives, they have limited capacity to deal with sparse inputs, ambiguity caused by occluded parts of the scene, and long-range extrapolations caused by drastic camera transformations~\cite{watson2022novel, chan2023generative}---all challenges that are particularly prominent in exo-to-ego view translation.  Alternatively, generative-based novel view synthesis methods~\cite{liu2021infinite, ren2022look, rombach2021geometry, wiles2020synsin, watson2022novel, tseng2023consistent, chan2023generative} use generative priors to synthesize random plausible outputs from the conditional distribution, making them adept at handling ambiguity and extrapolating to occluded parts of the scene.
Our Exo2Ego framework belongs to the generative-based vein, but unlike all the above, it does not depend on camera poses as critical conditional inputs. 
This improves scalability and widens the scope of potential applications, as rigid camera parameters are seldom accessible in real-world scenarios and are challenging to infer solely from RGB images.

\vspace*{-0.12in}
\paragraph{Cross-view Image-to-Image Translation}
Exo-to-ego view synthesis is inherently an image-to-image translation problem~\cite{isola2017image,wang2018high,wang2018video,bansal2018recycle}---but with drastic viewpoint change. 
Prior work centers around aerial-to-ground translation~\cite{REGMI2019, Regmi_2018_CVPR, tang2019selectiongan, ren2021crossmlp} or generating an egocentric view without human-object interactions, i.e. a person walking indoors~\cite{tang2019selectiongan, liu2020pgan, liu2021stagan}.
Except for P-GAN~\cite{liu2020pgan}, all of these approaches require the ground truth semantic map of the target domain at training or inference time to ensure high-quality generation---an unrealistic assumption that means a good portion of the translation task (the boundaries of the objects in the unobserved ego view or the input exo view) is already known, 
Compared with P-GAN, our Exo2Ego exhibits significantly better qualitative and quantitative results.

Diffusion Models (DMs) have shown competitive performance in image synthesis~\cite{ho2020denoising, song2021scorebased, dhariwal2021diffusion, ho2022cascaded}, but their use for cross-view translation remains unexplored.
We develop an exo-to-ego translation framework incorporating the diffusion model over GANs, considering potential advantages such as addressing mode collapse, promoting sample diversity, and maintaining stable training dynamics for improved image generation~\cite{rombach2022high}.

\section{Methodology}
First, we formalize our problem (Sec.~\ref{sec:formulation}).  Then we present the two stages of our model (Sec.~\ref{sec:hst} and~\ref{sec:plh}), followed by an overview of training and inference (Sec.~\ref{sec:training}).

\subsection{Problem Formulation}\label{sec:formulation}
Whereas 
conventional geometry-aware approaches~\cite{mildenhall2020nerf, Yu_2021_CVPR} 
generate novel views by 
specifying camera poses and performing volume rendering, 
Exo2Ego offers a different perspective. We propose tackling the exo-to-ego translation problem 
in a purely probabilistic framework. Intuitively, this is essential to address the inherent ambiguity in predicting the ego view from the exo view, e.g., due to entirely unseen portions of objects or the human body.

We define $\mathbf{X}_T=\{x_1,\ldots,x_i,\ldots, x_T\}$ as a series of $T$ video frames captured from an exo viewpoint, characterized by static backgrounds with dynamic actors and other objects present, where $i$ denotes the time index. This exo perspective reveals the actors' motions and (potentially) full-body poses within the scene. 
Let $\mathbf{Y}_T=\{y_1,\ldots,y_i,\ldots, y_T\}$ denote the corresponding sequence of ego frames captured from a first-person perspective. This perspective emulates the view of a camera mounted on the head or body 
of the actor, with a focus on their actions and interactions.

The goal of exo-to-ego view translation is to simulate the view of the ego camera-wearer in the scene recorded by an exo camera. Formally, we 
seek a translation model that can map $\mathbf{X}_T$ into a series of output exo frames, $\hat{\mathbf{Y}}_T =\{\hat{y}_1,\ldots,\hat{y}_i,\ldots,\hat{y}_T\}$. As shown in Eqn.~(\ref{eq:mapping}), the conditional distribution of $\hat{\mathbf{Y}}^T$ given $\mathbf{X}^T$ should be indistinguishable from the conditional distribution of $\mathbf{Y}^T$ given $\mathbf{X}^T$.
\begin{equation}
\label{eq:mapping}
    p(\hat{\mathbf{Y}}_{T}|\mathbf{X}_{T}) = p(\mathbf{Y}_T|\mathbf{X}_T).
\end{equation}
We prioritize translating daily tabletop activities from the exo to ego view, which is a prevalent setting in egocentric learning and frequently entails extensive interactions between hands and objects. This problem is highly challenging as the translation model must produce photorealistic hand-object interaction sequences in the ego view while also performing geometric and semantic reasoning, i.e., correctly predicting the spatial location and pixel-level appearance of visual concepts. 
Furthermore, it requires linking the actor's head/body motion captured by the exo camera with the viewpoint change in the ego video.
\label{sec:method}
Upon examining the application of the widely used pixel-to-pixel generation method~\cite{wang2018high} for our task, we found it struggles with intricate details of the hands, likely because treats all pixels equally and lacks geometric correspondence between views.  At the same time, we know that (at least in manipulation-rich scenarios) the hands of the actor are a common ground between the ego and exo views, despite their many other differences.

Motivated by these points, we propose the Exo2Ego framework, which disentangles the understanding of cross-view correspondences and pixel-level synthesis.
It consists of two key modules, as illustrated in Figure~\ref{fig:framework}:
(1) High-level Structure Transformation, which addresses the task of inferring the location and interaction manner of hands and objects in the ego view. To accomplish this, we train a transformer-based encoder-decoder model to translate the exo frame into an ego hand-object interaction layout (detailed below). 
(2) Diffusion-based Pixel Hallucination, which learns to synthesize realistic and high-quality pixel-level details by training a conditional diffusion model operating on top of the ego hand layout.

\subsection{High-level Structure Transformation}
\label{sec:hst}
Given an exo frame, the purpose of the high-level structure transformation is to train a layout translator which predicts the ego layout showing the location and rough contour of the visual concepts.  Specifically, to capture fine-grained hand-object interaction details, we propose to generate the hand layout, instantiated as 2D hand poses.
The quality of the generated layout is crucial, as it serves as a crucial reference for further pixel-level hallucination. 
To achieve this, we draw inspiration from a recent study~\cite{rombach2021geometry} that highlights transformer-based architectures~\cite{vaswani2017attention, dosovitskiy2020vit} succeed in understanding cross-view correspondence, due to their reduced locality-bias. 
Considering transformers have the potential to represent the geometric information implicitly, we employ a pure transformer-based encoder-decoder architecture, which enables us to effectively incorporate and process the comprehensive context of the exo scene, and thus help generate a precise ego hand layout.

\paragraph{Exo Contextual Feature Extraction} 
As shown in Figure~\ref{fig:framework}, an exo frame $x_i \in \mathbb{R}^{H \times W \times C_1}$ concatenated with the corresponding exo layout $x_i^l \in \mathbb{R}^{H \times W \times C_2}$ (2D hand pose layout recorded as image size) is first split into a sequence of patches and then processed by an input embedding layer, such as patch embedding for ViTs~\cite{dosovitskiy2020vit} to get a tokenized input sequence $e_i \in \mathbb{R}^{M \times D}$. To capture the positional information which is critical for spatial relationship reasoning, we add learnable position embedding to the sequence of patches. Then the sequence of tokens is passed into a transformer encoder consisting of $N$ repeated blocks. Each block contains a multi-head attention token mixer to first communicate information among tokens, which is represented as:
\begin{equation}
    z_i^{j-1} = \mathrm{Attention}(\mathrm{Norm}(e_i^{j-1})) + e_i^{j-1},
\end{equation}
where $j \in \{1, ..., N\}$ and $\mathrm{Norm}(\cdot)$ is the normalization method.
Then, the mixed token $z_i^{j-1}$ is passed into a two-layer MLP expressed as follows:
\begin{equation}
    e_i^j = \mathrm{MLP}(\mathrm{Norm}(z_i^{j-1})) + z_i^{j-1}.
\end{equation}

After iterating through all $N$ blocks, the input sequence $e_i$ undergoes a mapping process that transforms it into the final contextual tokens $c_i\in \mathbb{R}^{M \times D}$. We talk about how to decode the contextual tokens into the desired ego layout in the following part.

\paragraph{Layout Decoder} Consider a hand pose image with shape $\{H \times W \times 3\}$ 
as the target ego layout.  The contextual tokens $c_i$ should be decoded into 2D hand joint coordinates and then rendered to the RGB image. 
First, the contextual tokens $c_i$ are combined with a sequence of learnable query embeddings $q_i \in \mathbb{R}^{E \times D}$ and passed into a decoder architecture, consisting of $K$ transformer blocks, as described earlier. Following the transformer blocks, a regression head is employed to estimate the 2D coordinates of the joints whose range is restricted to $[0, 1]$.
Then we apply the classic bipartite matching loss~\cite{carion2020detr, redmon2016yolo} to map the regressed hand joints to the ground truth joints. 

\subsection{Diffusion-based Pixel Hallucination}
\label{sec:plh}
Having inferred the ego hand pose layout  $y_i^l$, next the aim of pixel-level hallucination is to synthesize photorealistic ego frames by taking into account an exo frame $x_i$ and the target ego layout.
We use the diffusion formulation proposed in the denoising diffusion probabilistic model (DDPM)~\cite{ho2020denoising} and train the diffusion model in the latent space.  As shown in Figure~\ref{fig:framework}, we first adopt 
a pre-trained variational autoencoder (VAE) model~\cite{kingma2013auto} as used in~\cite{rombach2022high} to encode the original ego frames $y_i$ and conditional information exo frame $x_i$ and ego layout $y_i^l$ into the latent space. 
Then, we train a diffusion transformer~\cite{Peebles2022DiT} to learn the latent data distribution by denoising a latent vector sampled from a Gaussian distribution gradually. 
Specifically, given an initial noise map $\bepsilon \sim \mathcal{N}(\bzero, \bI)$ and a conditioning vector $\bd$, the diffusion model generates the corresponding ego latent $\hat\bz_\theta$.
A squared error loss is used to denoise a variably-noised latent code $\bz^n = \alpha^n \bz + \sigma^n \bepsilon$ as follows:
\begin{equation}
\label{eq:diffusion}
    \mathcal{L} = \Eb{\bd, \bz,\bepsilon,n}{\|\bz - \hat\bz_\theta(\alpha^n \bz + \sigma^n \bepsilon, \bd) \|^2_2}
\end{equation}
where $\bz$ is the ground-truth latent vector, $\bd$ is the conditioning vector, and $\alpha^n, \sigma^n$ are functions of the diffusion process time step $n$, which control the noise schedule and sample quality.
We simply concatenate the noisy latent vector $\bz^n$ and the conditional embedding $\bd$ as the input of the denoising transformer. 
In contrast to earlier cross-view translation methods based on GANs~\cite{wang2018high, wang2018video, liu2020pgan}, our conditional diffusion model sequentially updates the target ego outputs, excels in capturing intricate ego-exo dependencies and faithfully reproducing the mapping of the conditional distributions in Eqn. (\ref{eq:mapping}). Notably, our model exhibits enhanced stability throughout the training and consistently produces higher-quality samples, as verified by our experiments.

Our diffusion model operates on a per-frame basis, independent of any previously generated ego frames or observed exo frames from the past. However, our Exo2Ego framework can be seamlessly integrated into video-to-video synthesis techniques, such as vid2vid~\cite{wang2018video}, to enhance the temporal coherence in the resulting videos. For example, Exo2Ego can generate the initial ego frame which could serve as the initialization of vid2vid’s sequential generation.

\subsection{Training and Inference}\label{sec:training}
At the training stage, the two modules described in Section~\ref{sec:hst} and ~\ref{sec:plh} are trained separately. During inference, we first generate the corresponding ego hand layout prediction $\hat{y}_i^l$ for each input exo frame $x_i$ with the layout translator described in Section~\ref{sec:hst}. Subsequently, $\hat{y}_i^l$ is concatenated with $x_i$ in a channel-wise manner and then used as the conditional input of the pixel-level diffusion generator described in Section~\ref{sec:plh}.

\section{Experimental Evaluation}
\label{exp}

We introduce an exo-to-ego synthesis benchmark (Sec.~\ref{sec:benchmark}) and then validate our approach against state-of-the-art synthesis methods (Sec.~\ref{sec:quality}).


\subsection{A Benchmark for Exo-to-Ego View Translation}\label{sec:benchmark}

To facilitate new work on the exo-to-ego translation task, we contribute a new cross-view synthesis benchmark sourced from three public time-synchronized multi-view datasets: H2O~\cite{Kwon_2021_ICCV}, Aria Pilot~\cite{aria_pilot_dataset}, and Assembly101~\cite{sener2022assembly101}. 
\textbf{H2O} \hoo ~\cite{Kwon_2021_ICCV} has indoor videos featuring actors manipulating 8 different objects using both hands. 
\textbf{Assembly101} \assembly ~\cite{sener2022assembly101} shows people (dis)assembling 101 take-apart toy vehicles. We take 6 activity sequences involving different individuals manipulating a toy roller.
\textbf{Aria Pilot} \aria ~\cite{aria_pilot_dataset} provides 16 multi-view recordings of an actor wearing Project Aria glasses~\cite{somasundaram2023projectaria} manipulating YCB~\cite{calli2015ycb} objects on a desktop. 

For all datasets, we select the exo camera viewpoints that offer the clearest capture of the action 
(`cam 2' for H2O, `214-7' for Aria Pilot, and `v4' for Assembly101).  
All frames are cropped and resized into $256\times256$. We divide each video into 30-frame clips.
Overall, this dataset showcases an array of tabletop activities with various objects, from toy assembly to the manipulation of everyday objects. More details are provided in Appendix.

We employ the following split settings to evaluate four kinds of generalization: 
(1) \textbf{new actions}, where we put the first 80\% of a video's clips into the training set and the remaining 20\% in the test set, hence splitting up action steps over time (e.g., for a video depicting picking up a coffee box and then taking coffee capsules out, the former will be assigned to the training set and the latter to the test set); (2) \textbf{new objects}, where we train with videos involving any of 7 objects and test with videos containing a novel 8th object; (3) \textbf{new subjects}, where we train with all clips from one subject and test on clips from another (subject 1 $\rightarrow$ subject 2); (4) \textbf{new scenes (backgrounds)}, where we train and test with disjoint scenes (scene h1, h2, k1, k2 $\rightarrow$ scene o1, o2).  
We analyze new action generalization on all 3 datasets, and 
the rest on H2O only, as it is the only dataset with the variations and meta-data to allow such split settings.
For action generalization, H2O, Aria Pilot, and Assembly101 have 704, 343, and 682 training clips, and 199, 95, and 205 test clips, respectively. For other generalization settings (new objects, subjects, and scenes), there are 691, 704, and 591 training clips, and 108, 194, and 312 test clips.


\begin{figure}[t]
\centering
\includegraphics[scale=0.43]{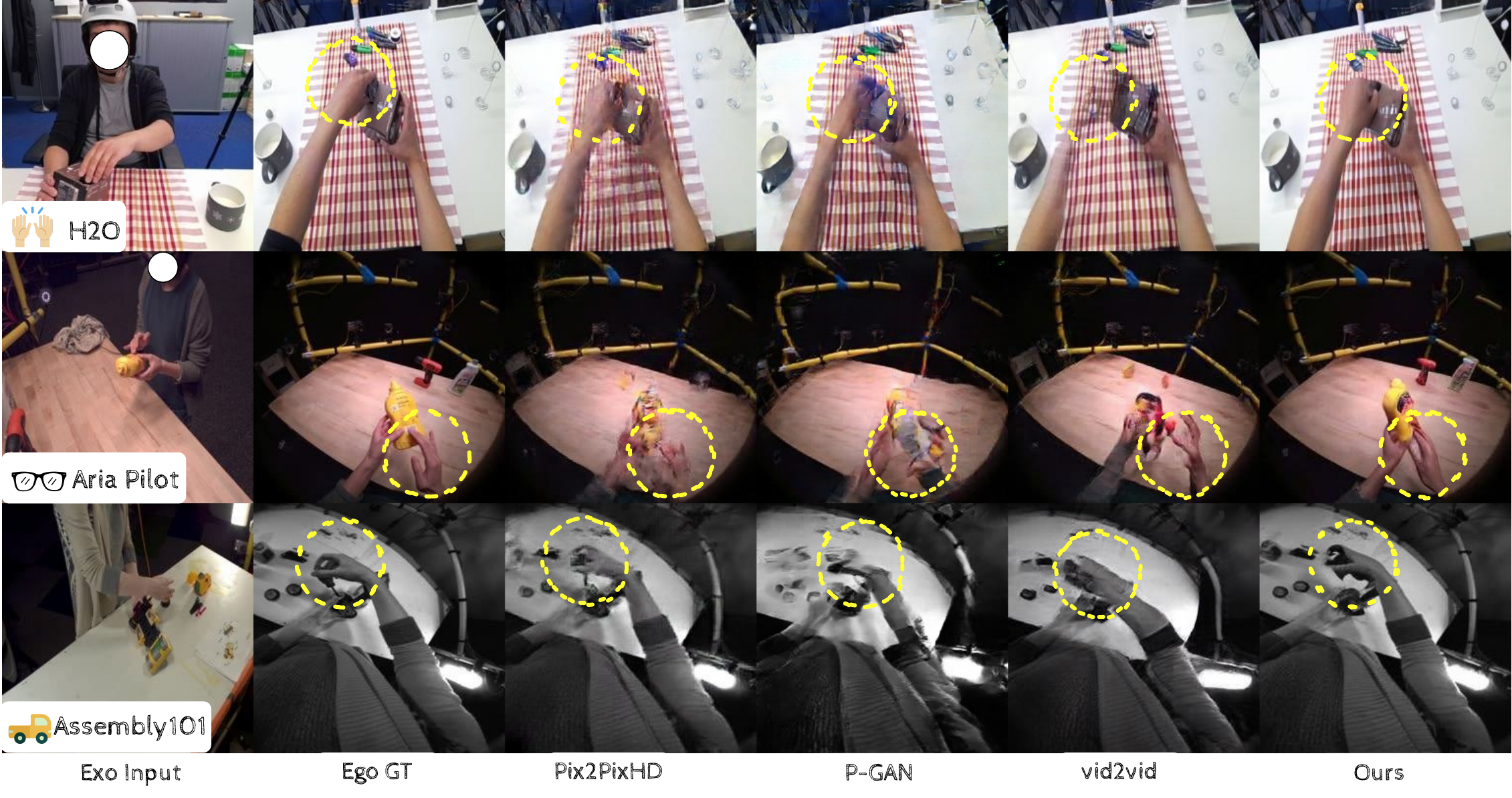}
\caption{Qualitative examples when generalizing to new actions on all datasets. More examples are in Appendix.}
\label{fig:main_vis}
\end{figure}

\begin{table}[t]
\renewcommand\thetable{1} 
  \centering
  \begin{tabular}{l|ccc cccc}
  
 & \multicolumn{7}{c}{H2O \hoo}\\
\thickhline
Method & SSIM$\uparrow$ & PSNR$\uparrow$ & FID$\downarrow$ & $P_\text{Squeeze}$$\downarrow$ &  $P_\text{Alex}$$\downarrow$ & $P_\text{Vgg}$$\downarrow$ & Feasi.$\uparrow$ \\

\hline
pix2pixHD~\cite{wang2018high}  & 0.428 & 30.370 & 132.03 & 0.163 & 0.220 & 0.337 & 0.8952\\  
P-GAN~\cite{liu2020pgan}   & 0.272 & 29.079 & 192.26 & 0.242 & 0.308 & 0.449 & 0.8353\\
vid2vid~\cite{wang2018video} & 0.402 & 29.882 & 85.76 & 0.166 & 0.226 & 0.342 & 0.9116  \\
pixelNeRF$^\ast$~\cite{Yu_2021_CVPR} & 0.219 & 27.871 & 470.13 & 0.711 & 0.807 & 0.731 & 0.0086 \\
\rowcolor{Gray}  Exo2Ego (Ours)  & \best{0.433} & \best{30.564} & \best{38.03} & \best{0.128} & \best{0.177} & \best{0.295} & \best{0.9758}\\
\hline
& \multicolumn{7}{c}{ Aria Pilot \aria  }\\
\thickhline
Method & SSIM$\uparrow$ & PSNR$\uparrow$ & FID$\downarrow$ & $P_\text{Squeeze}$$\downarrow$ &  $P_\text{Alex}$$\downarrow$ & $P_\text{Vgg}$$\downarrow$ & Feasi.$\uparrow$ \\

\hline
pix2pixHD~\cite{wang2018high} & 0.350 & 29.553 & 159.91 & 0.304 & 0.362 & 0.480 & 0.0790\\  
P-GAN~\cite{liu2020pgan} & 0.343 & 29.769 & 122.40 & 0.287 & 0.345 & 0.475 & 0.1014  \\
vid2vid~\cite{wang2018video} & 0.359 & 29.858 & 52.38 & 0.266 & 0.328 & 0.443 & 0.2646\\
\rowcolor{Gray}  Exo2Ego (Ours) & \best{0.371} & \best{29.952} & \best{26.01} & \best{0.245} & \best{0.305} & \best{0.421} & \best{0.5869}\\
\hline
 & \multicolumn{7}{c}{Assembly101 \assembly}\\
\thickhline
Method & SSIM$\uparrow$ & PSNR$\uparrow$ & FID$\downarrow$ & $P_\text{Squeeze}$$\downarrow$ &  $P_\text{Alex}$$\downarrow$ & $P_\text{Vgg}$$\downarrow$ & Feasi.$\uparrow$ \\

\hline
pix2pixHD~\cite{wang2018high} & 0.405 & 29.909 & 112.71 & 0.216 & 0.295 & 0.401 & 0.1184 \\  
P-GAN~\cite{liu2020pgan} & 0.372 & 29.303 & 114.68 & 0.216 & 0.300 & 0.436 & 0.0800\\
vid2vid~\cite{wang2018video} & 0.368 & 29.481 & 78.47 & 0.224 & 0.312 & 0.424 & 0.1358 \\
\rowcolor{Gray} Exo2Ego (Ours) & \best{0.406} & \best{30.027} & \best{33.16} & \best{0.178} & \best{0.254} & \best{0.365} & \best{0.3791}\\
\hline
\end{tabular}
  \caption{Generalizing to new actions on three datasets. $^\ast$pixelNeRF requires privileged camera information relative to the other baseline models. 
 }
    \label{tab:action}
\end{table}

\subsection{Evaluating Synthesis Quality}\label{sec:quality}
With this well-constructed benchmark at hand, we proceed to validate our Exo2Ego alongside baseline models.

\custompar{Baselines} 
We consider several baselines: (1) \texttt{Pix2PixHD}~\cite{wang2018high}, a single-view image translation method that processes the videos frame by frame; 
(2)  \texttt{P-GAN}~\cite{liu2020pgan}, a recent exo-to-ego view translation method that proposes a parallel generative network to facilitate cross-view image translation; 
(3)  \texttt{Vid2Vid}~\cite{wang2018video}, a single-view video translation method that models the temporal dynamics in the video;
(4) \texttt{pixelNeRF}~\cite{Yu_2021_CVPR}, a NeRF-like model known for its remarkable generalization capabilities and its accommodating stance toward sparser input views.
Note that pixelNeRF assumes camera parameters are known, while other methods do not. We conduct experiments with pixelNeRF only on H2O, as the other two datasets lack the camera information needed by pixelNeRF.

\begin{figure*}[t]
\centering
\includegraphics[scale=0.32]{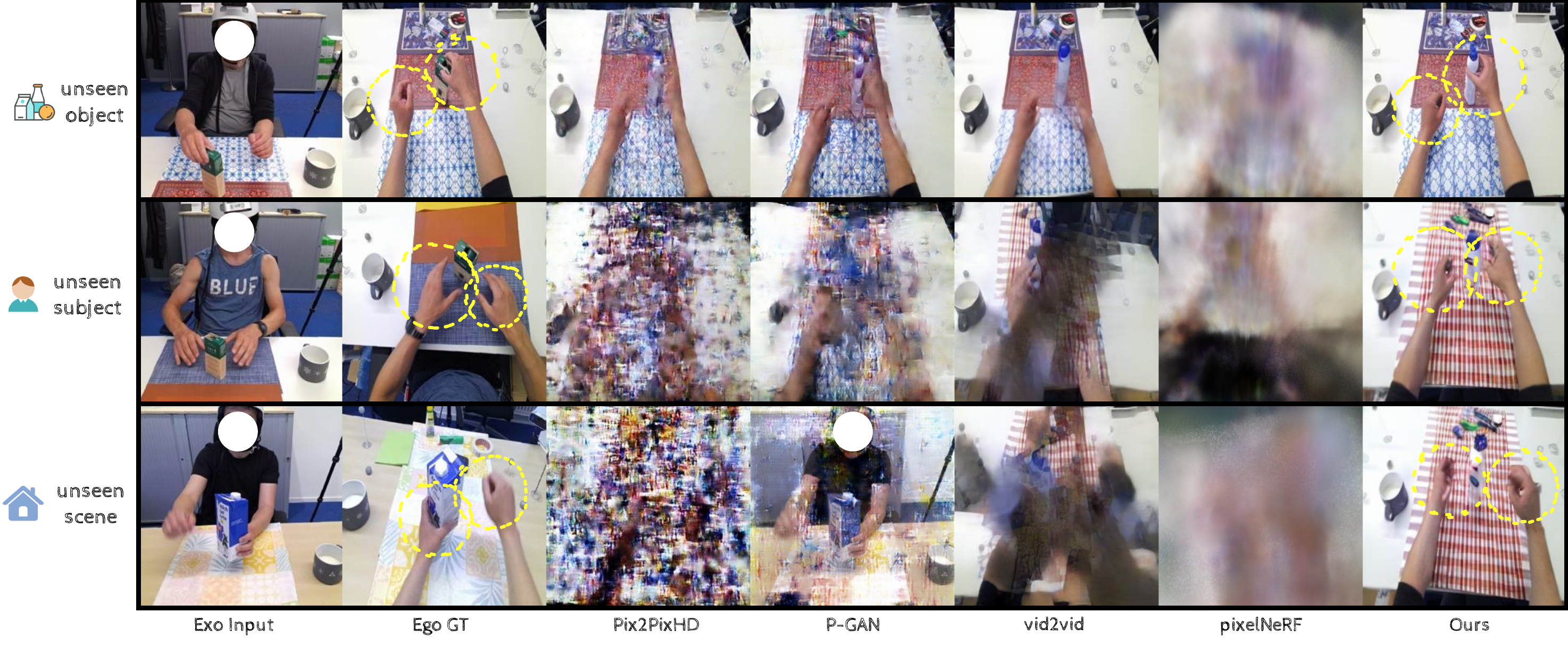}
\caption{Qualitative examples when generalizing to new objects, subjects, and scenes (backgrounds) on H2O dataset. 
}
\label{fig:gen_others}
\end{figure*}

\custompar{Implementation Details}
For Exo2Ego's high-level structure transformation transformer, the number of blocks $N$ is set to 6. We use DiT-XL/2 ~\cite{Peebles2022DiT} as the denoising architecture for Exo2Ego. Our diffusion model is trained for 40,000 steps for all datasets. 
pixelNeRF is trained for 70,000 steps for all datasets. 
For H2O and Assembly101, pix2pixHD, and P-GAN are trained for 100 epochs, and vid2vid is trained for 40 epochs. For Aria Pilot, we use 400 and 100 epochs respectively.  More details are in Appendix. 

\custompar{Evaluation Metrics} 
We measure the quality of the synthesized videos as follows.
First, we evaluate the feasibility of generated hands, denoted as \texttt{Feasi}. We adopt an off-the-shelf egocentric hand object detector~\cite{Shan_2020_CVPR} to measure the physical feasibility of the synthesized hands. This metric is calculated as the confidence score of hand detection results produced by the detector, averaged over all synthesized frames. The purpose of this metric is to check whether the synthesized hands are realistic enough to be faithfully detected. It’s noteworthy that~\cite{ye2023affordance} also employs the confidence score of the hand as an evaluation metric, but they focus on in-contact confidence rather than the hand detection confidence utilized in our approach. 
We also use structural similarity (\texttt{SSIM}) and peak signal-to-noise ratio (\texttt{PSNR})~\cite{hore2010SSIM} to check the pixel-level similarity between the predicted ego frame and the ground truth (GT) ego frame.
The evaluation metrics should ideally remain stable under minor transformations (such as small translations or affine transformations). So we further adopt perceptual metrics, including \texttt{FID}~\cite{Seitzer2020FID} and Learned Perceptual Image Patch Similarity (\texttt{LPIPS})~
\cite{zhang2018lpips} which evaluates the feature-level similarity of the predicted ego frame and the ground truth (GT) ego frame. LPIPS are denoted as $P_\text{Squeeze}$,  $P_\text{Alex}$, and $P_\text{Vgg}$, which uses SqueezeNet~\cite{iandola2016squeezenet}, AlexNet~\cite{krizhevsky2017alexnet}, and VGG~\cite{simonyan2014vgg} as the feature extractor, respectively.

\paragraph{Generalization to New Actions} 
Figure~\ref{fig:main_vis} showcases qualitative results on all datasets, demonstrating the superiority of our Exo2Ego framework. Compared to all baselines, our approach produces realistic hands with correct poses, especially noticeable in the highlighted yellow circle regions when dealing with new actions during test time. 
Table~\ref{tab:action} presents the feasibility of synthesized hands and the various measures of pixel-level similarity and perceptual similarity between the predicted ego frame and the ground truth ego frame for all datasets.
Our Exo2Ego greatly increases the feasibility of synthesized hands, with a confidence gain up to 32.23\% on Aria Pilot, compared with the best score of 26.46\% achieved by vid2vid. 
This improvement indicates enhanced realism in the generated hands, resulting in greater confidence in hand detection results. Additionally, our Exo2Ego can produce ego frames with higher pixel-level similarity to the ground truth frames and better perceptual metrics (much lower FID scores). This highlights the benefits of the `decoupling' idea of our Exo2Ego framework in generating photorealistic ego frames.


\begin{table}[t]
\renewcommand\thetable{2} 
  \centering
  \captionsetup{justification=centering} 
  \resizebox{1\linewidth}{!}{
  \begin{tabular}{l|ccc cccc}
  
 & \multicolumn{7}{c}{H2O-unseen object \object}\\
\thickhline
Method & SSIM$\uparrow$ & PSNR$\uparrow$ & FID$\downarrow$ & $P_\text{Squeeze}$$\downarrow$ &  $P_\text{Alex}$$\downarrow$ & $P_\text{Vgg}$$\downarrow$ & Feasi.$\uparrow$ \\

\hline
pix2pixHD~\cite{wang2018high}  & 0.333 & 29.504 & 253.73 & 0.269 & 0.360 & 0.467  & 0.7527 \\  
P-GAN~\cite{liu2020pgan}    & 0.270 & 28.915 & 264.76 & 0.277 & 0.350 & 0.492 & 0.7720\\
vid2vid~\cite{wang2018video}  & \best{0.349} & 29.585 & 145.39 & 0.242 & 0.324 & 0.428 & 0.8612 \\
pixelNeRF~\cite{Yu_2021_CVPR} & 0.262 & 27.912 & 473.73 & 0.750 & 0.841 & 0.736 & 0.0000 \\
\rowcolor{Gray}  Exo2Ego (Ours) & 0.332	& \best{29.701} & \best{102.58} & \best{0.228} & \best{0.308} & \best{0.419} & \best{0.9876} \\
\hline
& \multicolumn{7}{c}{H2O-unseen subject \subject}\\
\thickhline
Method & SSIM$\uparrow$ & PSNR$\uparrow$ & FID$\downarrow$ & $P_\text{Squeeze}$$\downarrow$ &  $P_\text{Alex}$$\downarrow$ & $P_\text{Vgg}$$\downarrow$ & Feasi.$\uparrow$ \\

\hline
pix2pixHD~\cite{wang2018high} & 0.069 & 28.028 & 477.47 & 0.661 & 0.754 & 0.765 & 0.0140\\  
P-GAN~\cite{liu2020pgan} & 0.088 & 28.063 & 451.37 & 0.623 & 0.708 & 0.725 & 0.2002 \\
vid2vid~\cite{wang2018video} & 0.136 & 28.094 & 308.52 & 0.564 & 0.690 & 0.686 & 0.3838 \\
pixelNeRF~\cite{Yu_2021_CVPR} & 0.170 & 27.874 & 459.49 & 0.769 & 0.836 & 0.768 & 0.0000 \\
\rowcolor{Gray}  Exo2Ego (Ours) & \best{0.178} & \best{28.266} & \best{150.87} & \best{0.503} & \best{0.629} & \best{0.665} & \best{0.9666} \\
\hline
 & \multicolumn{7}{c}{H2O-unseen scene \scene}\\
\thickhline
Method & SSIM$\uparrow$ & PSNR$\uparrow$ & FID$\downarrow$ & $P_\text{Squeeze}$$\downarrow$ &  $P_\text{Alex}$$\downarrow$ & $P_\text{Vgg}$$\downarrow$ & Feasi.$\uparrow$ \\

\hline
pix2pixHD~\cite{wang2018high}   & 0.024	& 27.918	& 558.87	& 0.729	& 0.789	& 0.788	& 0.0013\\  
P-GAN~\cite{liu2020pgan}  & 0.085	& 27.928	& 325.34	& 0.608	& \best{0.699}	& 0.724	& 0.6041 \\
vid2vid~\cite{wang2018video}  & 0.033	& 27.942	& 398.78	& 0.636	& 0.731	& 0.734	& 0.2095 \\
pixelNeRF~\cite{Yu_2021_CVPR} & 0.103 & 27.810 & 506.10 & 0.773 & 0.908 & 0.800 & 0.0000\\
\rowcolor{Gray}  Exo2Ego (Ours) & \best{0.157}	& \best{27.943}  & \best{249.04}	& \best{0.531}	& 0.709	& \best{0.658}	& \best{0.9635} \\
\hline
\end{tabular}}
  \caption{Generalizing to different objects, subjects, and scenes.}
    \label{tab:h2o}
\end{table}

\paragraph{Generalization to New Objects, Subjects, and Backgrounds} 
Illustrated in Table~\ref{tab:h2o}, our Exo2Ego outperforms the baselines in terms of quantitative metrics for generalization to new objects/subjects/backgrounds.
Figure~\ref{fig:gen_others} shows that our Exo2Ego can produce realistic hand-object interactions when encountering exo views with new distributions, while other baselines fail miserably. This is mainly because our diffusion-based generative mechanism gradually denoises the target ego view, capturing the inherent intricate mapping from exo to ego views better.
Another observation is that pixelNeRF produces ego views characterized by noticeable blurring and a lack of fine details in hands and objects. This observation supports our earlier discussions on the limitations of geometry-aware synthesis approaches.
\begin{figure*}[htbp!]
\centering
\captionsetup{justification=centering} 
\includegraphics[scale=0.32]{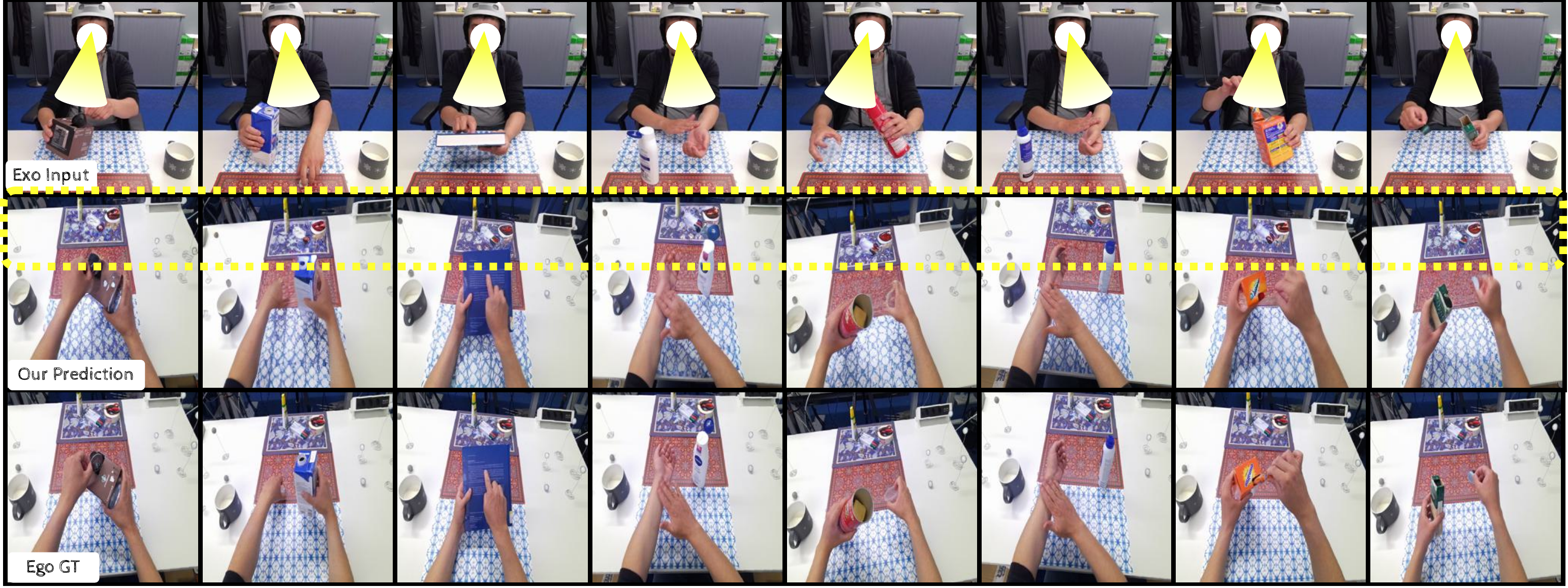}
\caption{Exo2Ego framework generates ego videos with reasonable viewpoint changes.}
\label{fig:head_motion}
\end{figure*}

\begin{figure*}[t]
\centering
\includegraphics[scale=0.28]{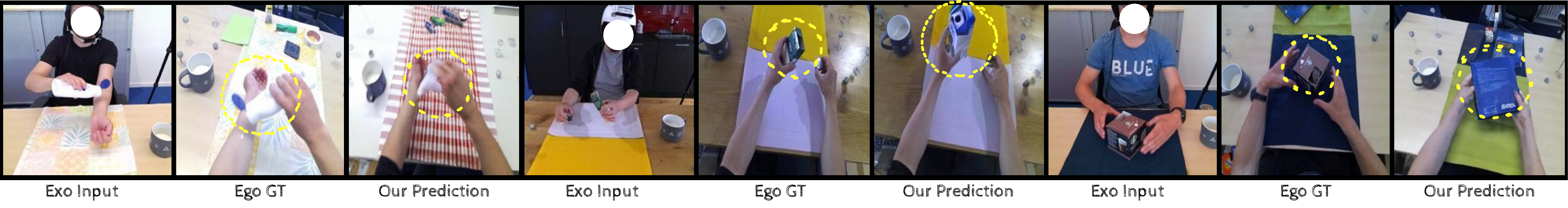}
\caption{Failure cases: Exo2Ego generates hands with reasonably accurate positions/poses but incomplete/wrong objects in the ego view.}
\label{fig:failure_case}
\end{figure*}

\paragraph{Modeling Egocentric Viewpoint Change}
Due to the probabilistic nature of our Exo2Ego framework, it excels in capturing the relationship between the viewpoint changes in the ego view and the head motions observed in the exo view, holding the potential to deliver more immersive virtual experiences. Intrinsically, our Exo2Ego framework leverages the fact that ego-camera pose is conditioned on the reference exo view, as the ego-camera wearer is usually visible in the exo view. 
This is highlighted by the yellow gaze direction and dashed box in Figure~\ref{fig:head_motion}. As the actor moves their head when interacting with objects, the slope of the desk edge in the ego view also changes, indicating a corresponding change in ego viewpoint. 
This implies the ego-camera motion can be implicitly inferred via conditional generative modeling, even if it's not enforced explicitly. 
The ability to model ego viewpoint changes is crucial, especially in physical activities like playing ball games which involve heavy head and body motions.

\section{Limitations and Future Work}
As an early-stage effort for exo-to-ego cross-view translation, our work makes non-trivial progress in producing reasonable hand manipulation details for novel actions. However, despite establishing stronger generalizations than the existing models, our model does not generalize perfectly to in-the-wild objects, subjects, and backgrounds, due to the modest scale of available training data.
Moreover, our qualitative results expose limitations in existing baselines and our Exo2Ego framework in generating 3D-consistent views for objects during test time, attributed to the absence of geometric priors for common objects, see Figure~\ref{fig:failure_case}.  Future work involves integrating robust object geometric priors. 
While leveraging predicted camera parameters for encoding 3D geometry is a promising approach, as shown Figure~\ref{fig:gen_others}, the popular 3D-aware method pixelNeRF~\cite{Yu_2021_CVPR} faces challenges in the dynamic exo-ego setting characterized by hand and object interactions, due to limitations in handling sparse inputs, occluded scenes, and long-range camera extrapolations. Notably, our Exo2Ego outperforms pixelNeRF, highlighting the potential for future work to explore explicit geometric reasoning within our framework.
We focus on hand-object interactions due to their significant value for applications in augmented reality and robotics, yet more general ego-exo settings will be interesting.

\section{Conclusion}
\label{sec:conclusion}
Overall, our work contributes to the growing body of research in cross-view translation and lays the groundwork for the important case of exo-to-ego. Our generative framework Exo2Ego integrates high-level structure transformation and pixel-level hallucination to yield very encouraging experimental results.   We also provide a curated benchmark task for supporting continued work on this problem. The core technology holds immense potential for applications in robot learning and AR skill coaching, where an ego actor needs to replicate the actions of a demonstrator observed from the exo perspective. 

\section{Acknowledgment}
\label{sec:acknowledgment}
This research has been supported by NSF Grants AF 1901292, CNS 2148141, Tripods CCF 1934932, IFML CCF 2019844 and research gifts by Western Digital, Amazon, WNCG IAP, UT Austin Machine Learning Lab (MLL), Cisco and the Stanly P. Finch Centennial Professorship in Engineering. UT Austin is supported in part
by the IFML NSF AI Institute. K.G. is paid as a research scientist at Meta.

\clearpage
\newpage
\bibliographystyle{assets/plainnat}
\bibliography{paper}

\clearpage
\newpage
\beginappendix
\appendix
\normalsize
\section{Experimental Setup}
\begin{figure*}[htbp]
    \centering
    \captionsetup{justification=centering} 
\includegraphics[scale=0.69]{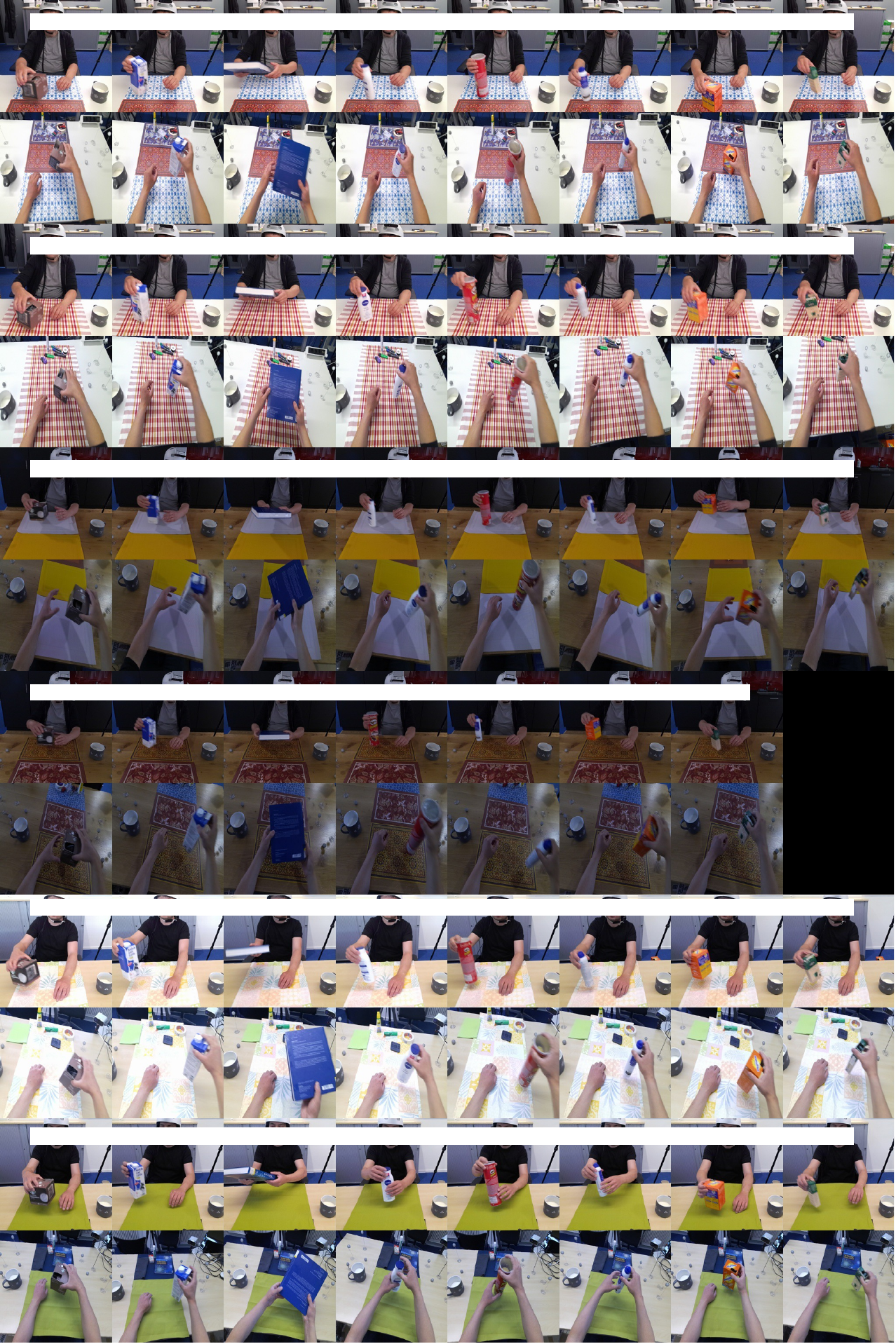}
\caption{Selected samples from the H2O dataset. 
}
\label{fig:dataset_h2o}
\end{figure*}

\begin{figure*}[htbp]
    \centering
    \captionsetup{justification=centering} 
\includegraphics[scale=0.40]{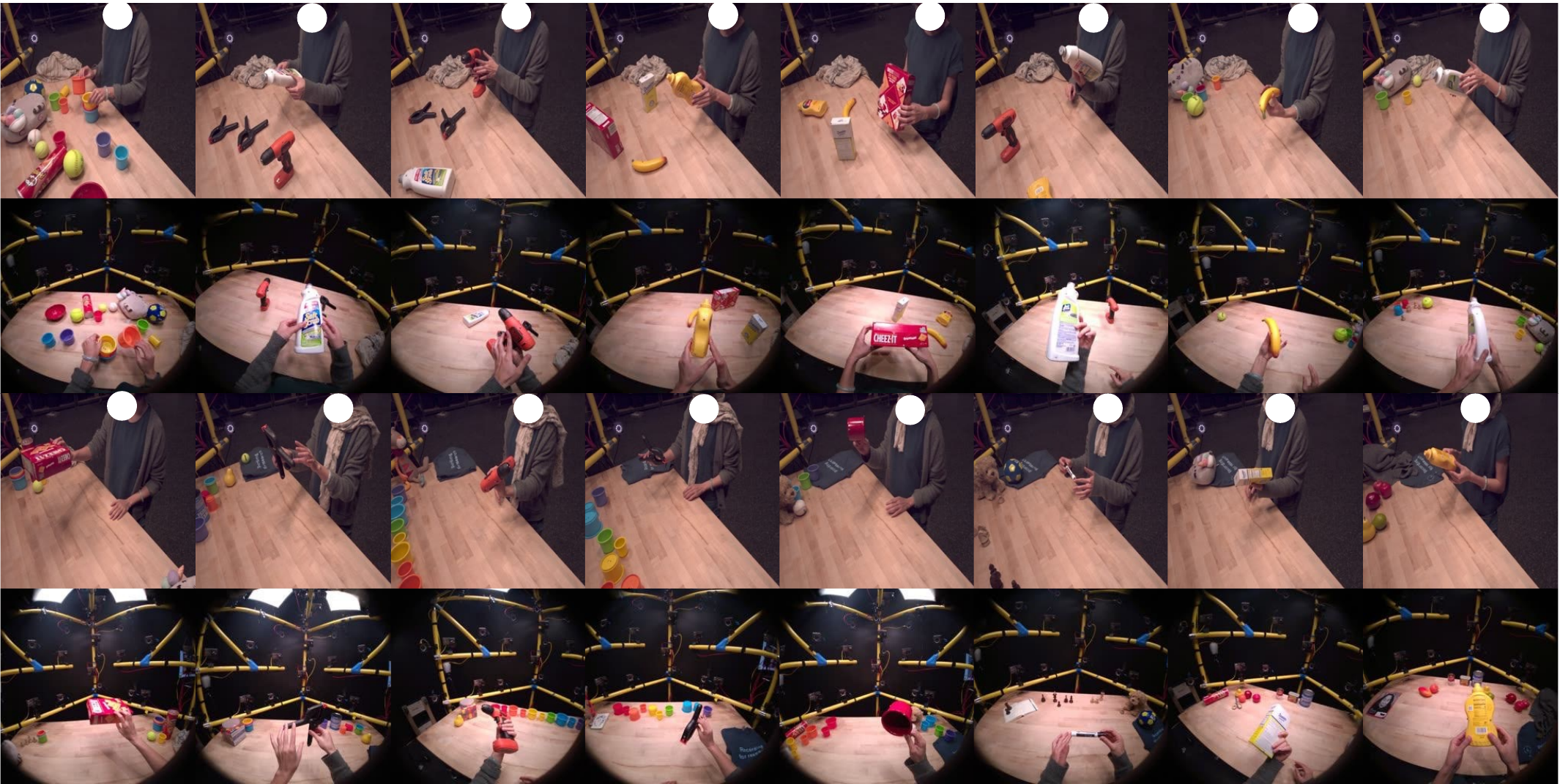}
\caption{Selected samples from the Aria Pilot dataset.}
\label{fig:dataset_aria}
\end{figure*}

\begin{figure*}[htbp]
    \centering
    \captionsetup{justification=centering} 
\includegraphics[scale=0.42]{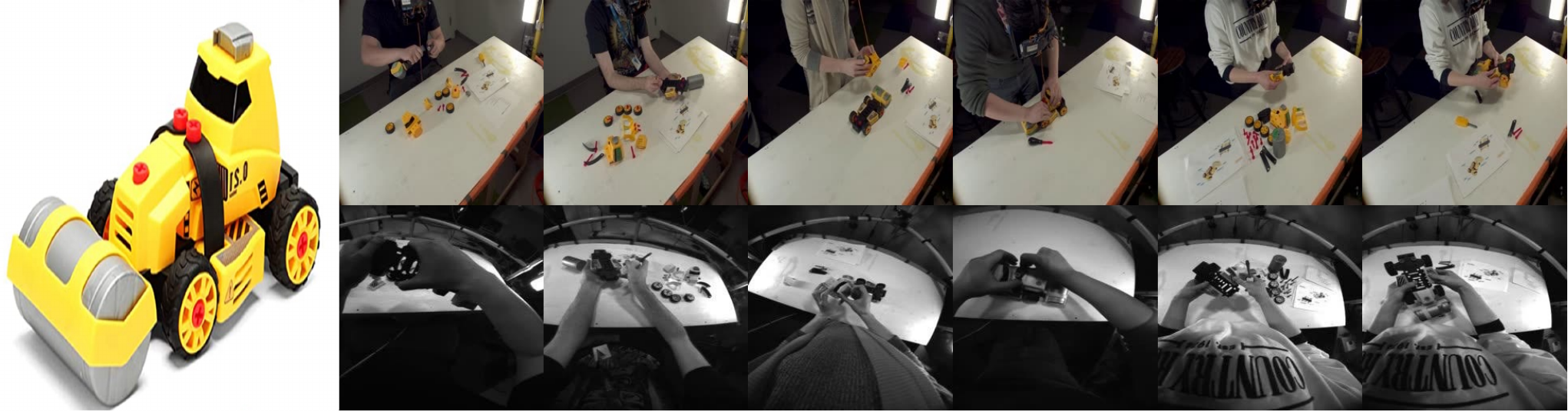}
\caption{Visualization of the toy roller and selected samples from the Assembly101 dataset.}
\label{fig:dataset_assembly}
\end{figure*}

\paragraph{Dataset Details}
Figure~\ref{fig:dataset_h2o},~\ref{fig:dataset_aria}, and~\ref{fig:dataset_assembly} show the preprocessed exo-ego frame pairs selected from H2O, Aria Pilot and Assembly101 respectively. For all the experiments, we perform frame cropping manually per scene to eliminate unnecessary background details and ensure that the actor is the focal point of each video. 
In the case of H2O, we focus on the videos of 'subject 1' and 'subject 2', in which the actors engage with 8 distinct everyday objects across 6 unique scenarios. We blurred human faces in the H2O dataset.
Regarding Aria Pilot, we focus on the desktop activities set. We exclude the frames in which the actor is not wearing the glasses.
These desktop activities involve tasks such as tidying up the desk, manipulating multiple objects, and manipulating single objects. 
The objects were primarily selected from the YCB Benchmark object set~\cite{calli2015ycb}, consisting of 10 commonly used items. Human faces are blurred in the Aria Pilot dataset.
For Assembly101, we select 6 sequences that depict 5 different actors assembling a toy roller. The rationale behind our downselection lies in the visual homogeneity identified in Assembly101's video content. In essence, we aim to guarantee that the selected sequences are representative of the overall dataset.
Note that Assembly101 has 4 ego views. We select 'e3' as the target ego view since it shows the clearest hand-object interaction details. 

\begin{figure*}[htbp]
    \centering
    \captionsetup{justification=centering} 
\includegraphics[scale=0.49]{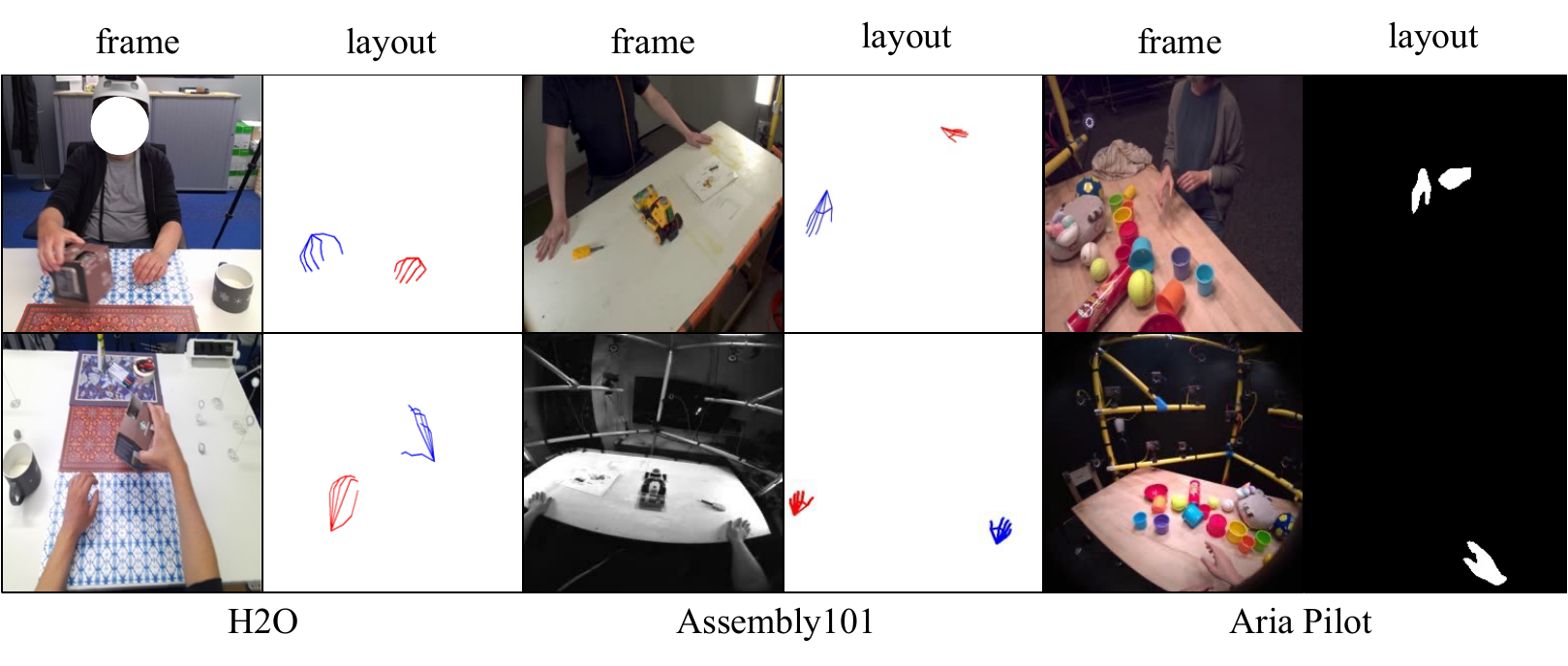}
\caption{Layout visualizations for all datasets.}
\label{fig:layout}
\end{figure*}

\paragraph{Layout Annotations}
As shown in Figure~\ref{fig:layout}, in our Exo2Ego framework, we utilize the 2D hand pose as the layout for H2O and Assembly101. To obtain the 2D hand poses, we project them from the 3D hand poses available in the original dataset.
In the case of Aria Pilot, it does not have 3D hand pose annotations, so we employ the hand mask as the layout. We first utilize an off-the-shelf hand detector~\cite{Shan_2020_CVPR} to obtain the 2D hand bounding boxes. Subsequently, we apply Segment Anything~\cite{kirillov2023sam} to generate the hand masks with the bounding boxes.

\paragraph{Implementation Details}
For all baselines except PixelNeRF, we utilize the Adam optimizer~\cite{kingma2014adam} with a learning rate of 0.0002 and betas set to (0.5, 0.999). For PixelNeRF and Exo2Ego, we use the Adam optimizer with a learning rate of 0.0001.
Regarding Exo2Ego's high-level structure transformation module, we adopt DeiT~\cite{touvron2021deit} as the transformer backbone.
For all methods, we monitor the training loss curve, ensuring that the losses exhibit a satisfactory convergence by the specified final epochs. Additionally, an evaluation of the synthesis results on the training set is performed to confirm the stability of synthesis quality and its tendency to plateau — further improvements are unlikely.
All experiments were conducted using PyTorch 1.10~\cite{paszke2019pytorch}.

\section{Mask as Layout}
In the hand mask layout case, we employ a hand mask with dimensions of $\{H \times W \times 1\}$  as the target ego layout. The overall architectures of the transformer encoder and decoder remain consistent with the hand pose case. The only distinction lies in the decoder, which produces a feature map with dimensions of $\{H \times W \times 2\}$ rather than hand joint locations. Treating mask generation as a pixel-wise classification problem (hand region vs. non-hand region), we define the number of classes to be 2. For the decoder output, we apply softmax to the class dimension to derive the final mask prediction. The model is trained end-to-end using per-pixel cross-entropy loss.

\section{Extra Experimental Results}
\begin{figure*}[htbp]
\centering
\captionsetup{justification=centering} 
\includegraphics[scale=0.67]{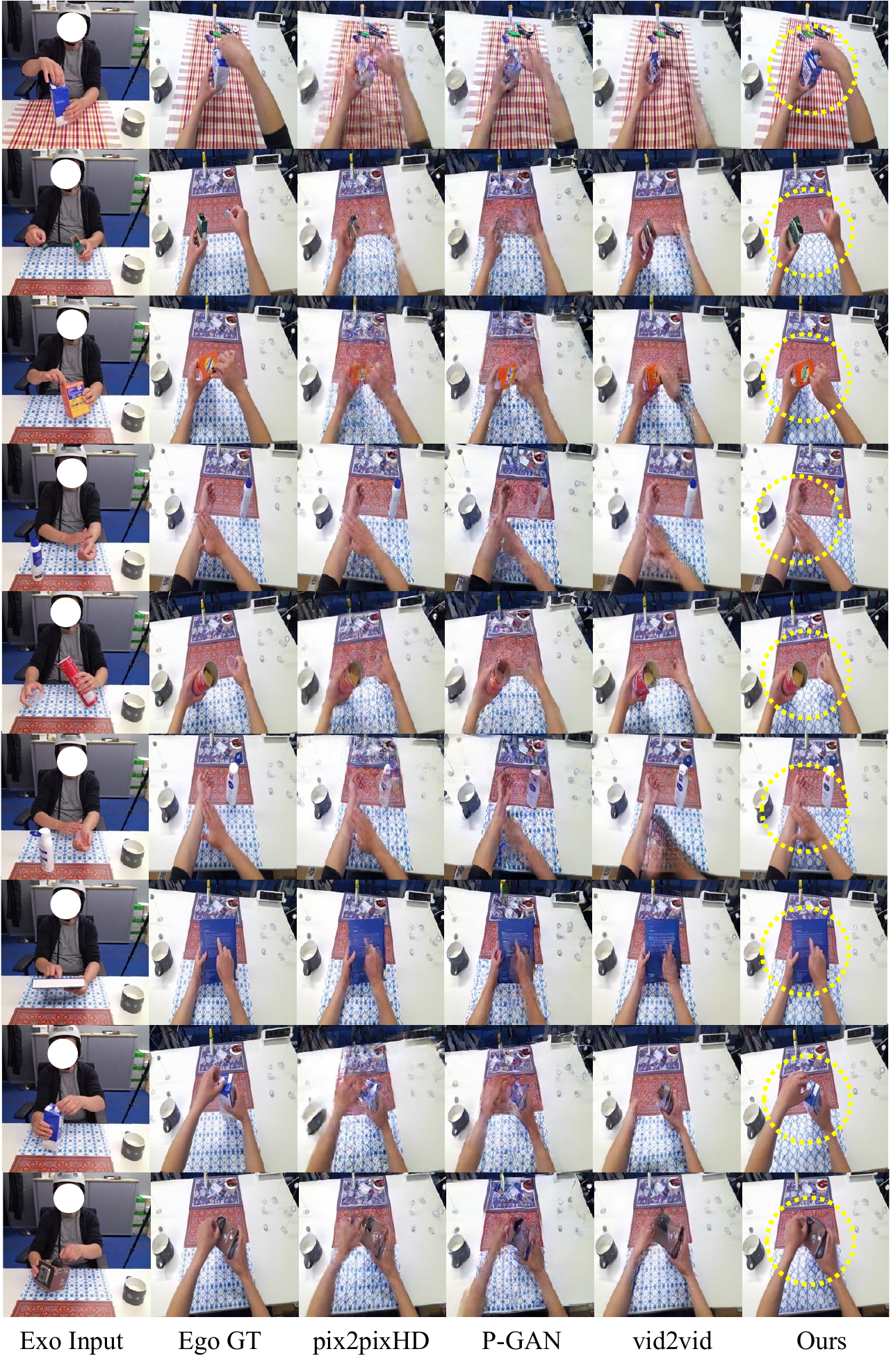}
\caption{Extra qualitative results of generalizing to new actions on H2O.}
\label{fig:h2o_results}
\end{figure*}

\begin{figure*}[htbp]
\centering
\captionsetup{justification=centering} 
\includegraphics[scale=0.67]{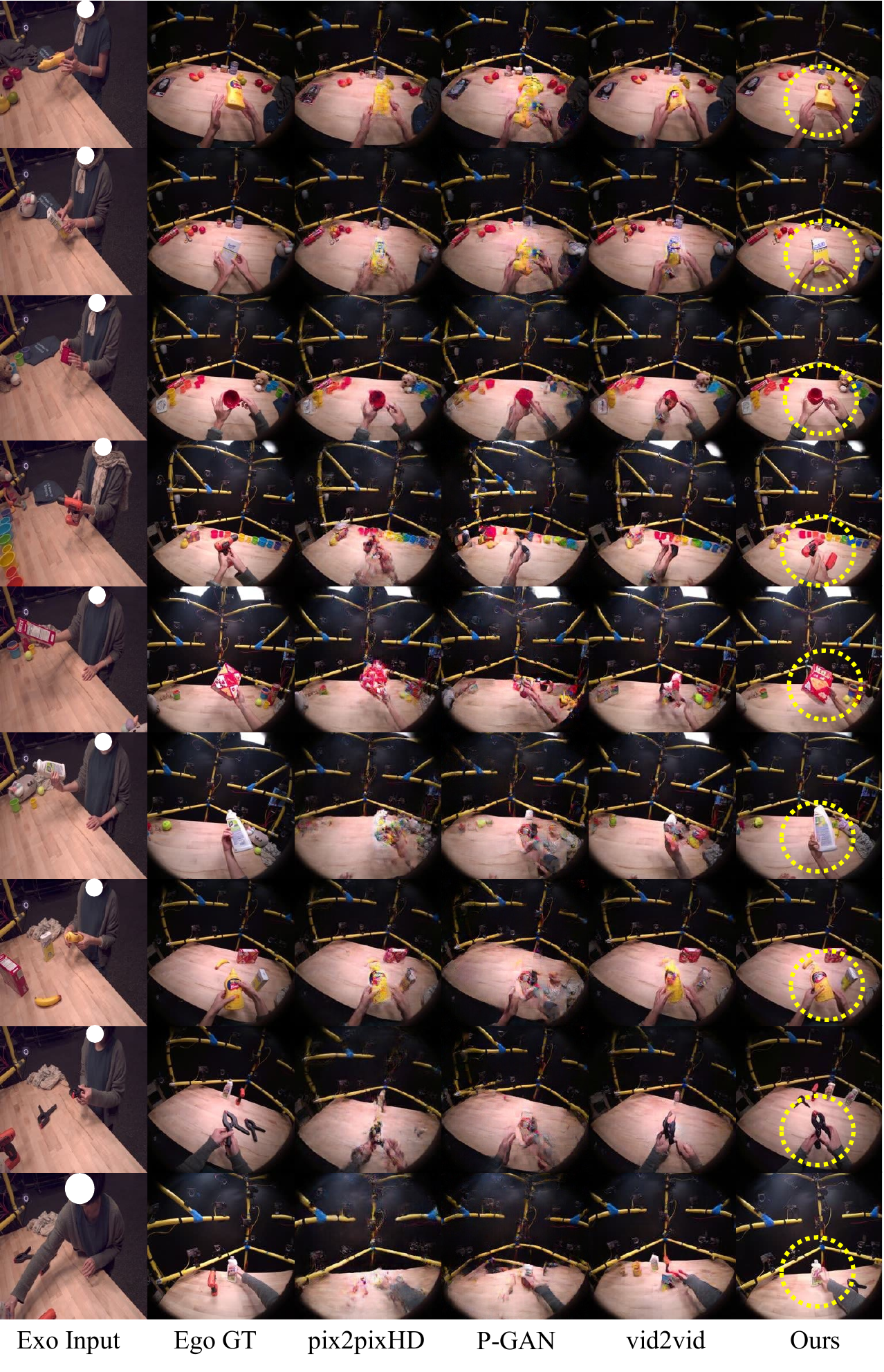}
\caption{Extra qualitative results of generalizing to new actions on Aria Pilot.}
\label{fig:aria_results}
\end{figure*}

\begin{figure*}[htbp]
\centering
\captionsetup{justification=centering} 
\includegraphics[scale=0.68]{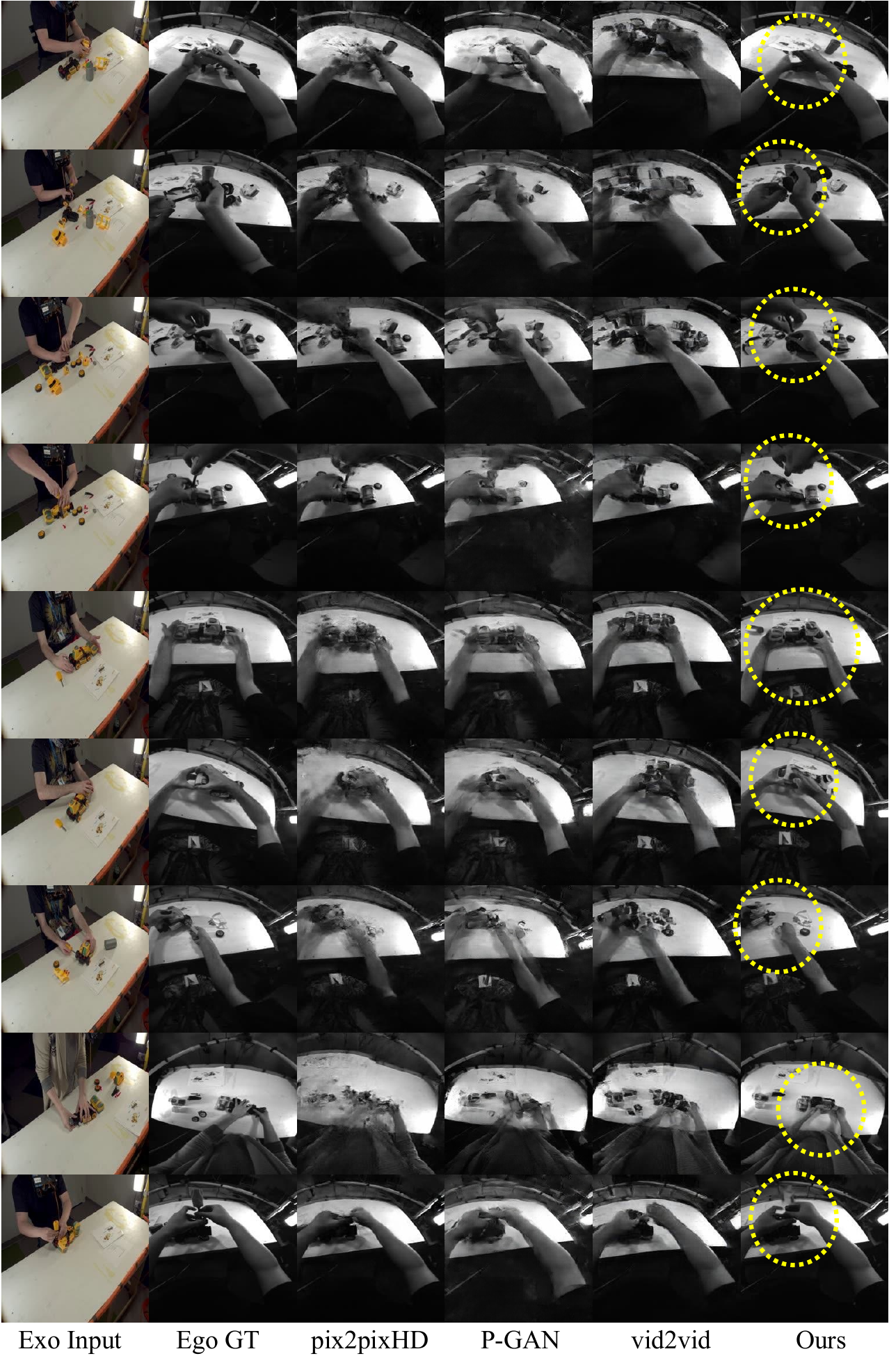}
\caption{Extra qualitative results of generalizing to new actions on Assembly101.}
\label{fig:assembly_results}
\end{figure*}

\paragraph{Additional Qualitative Results} We offer additional qualitative results of generalizing to new actions on all datasets in Figure~\ref{fig:h2o_results},~\ref{fig:aria_results}, and~\ref{fig:assembly_results}. Our Exo2Ego framework has demonstrated superior performance compared to all baseline methods when it comes to synthesizing hand-object interactions that are realistic and visually coherent. 

\begin{figure}[t]
    \centering
    \captionsetup{justification=centering} 
\includegraphics[scale=0.17]{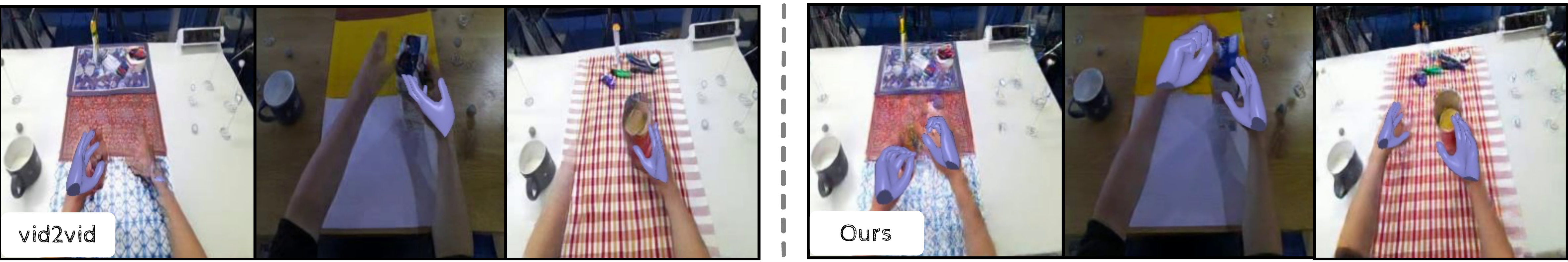}
\caption{Visualizing mesh rendered by a pretrained 3D hand estimation model.}
\label{fig:hand_rendered}
\end{figure}

\paragraph{Ablation Study on the Choice of Source Exo View}
We simply choose the exo view with the largest FOV of the ego view – showcasing the intricate details of hand and object interaction most effectively. 
We have conducted ablation studies on the selection of different exo views (cam 0-4) within the H2O dataset. As shown in Table~\ref{tab:rebuttal_1}, we observe that different sources for the exo view do impact the model's performance, and the exo view – cam 2, chosen by us – with the highest FOV alongside the ego view, delivers the best translation performance.

\begin{table}[t]
\renewcommand\thetable{2} 
    \centering
    \captionsetup{justification=centering} 
  \caption{Using different exo cameras as source views on H2O.}
  \vspace{5pt}
  \label{tab:rebuttal_1}
  \resizebox{0.58\linewidth}{!}{
  \begin{tabular}{l|ccc ccc}
    \makecell[l]{Source \\Camera} & SSIM$\uparrow$ & PSNR$\uparrow$ & FID$\downarrow$ & $P_\text{Squeeze}$$\downarrow$ & $P_\text{Alex}$$\downarrow$ & $P_\text{Vgg}$$\downarrow$ \\
    \thickhline
    cam 0 & 0.427 & \best{30.394} & 135.74 & 0.165 & 0.222 & 0.338\\  
    cam 1 & 0.416 & 30.226 & 142.59 & 0.172 & 0.231 & 0.347 \\
    \rowcolor{Gray} cam 2 & \best{0.428} & 30.370 & \best{132.03} & \best{0.163} & \best{0.220} & \best{0.337} \\
    cam 3 & 0.410 & 30.281 & 143.21 & 0.184 & 0.243 & 0.356 \\
\hline
\end{tabular}}
\end{table}

\paragraph{Extracting 3D Hand Pose}
Following~\cite{ye2023affordance}, we employ FrankMocap~\cite{rong2021frankmocap, joo2020eft}, an off-the-shelf hand pose estimator, to directly extract 3D poses from the generated ego frames. The extracted 3D poses are visualized in Figure~\ref{fig:hand_rendered}. It is observed that our Exo2Ego framework surpasses vid2vid in synthesizing highly realistic and visually coherent hands, enabling the rendering of exceptional 3D hand meshes that exhibit accurate positioning and overall realism. Note that the quality of 3D hand mesh is essential for creating vivid and lifelike experiences in virtual and augmented reality (VR/AR).

\section{Limitations and Future Work}
\paragraph{Generalization Ability}
There are different types of generalization. As an early-stage work for exo-to-ego cross-view translation, our paper has made non-trivial progress in producing reasonable hand manipulation details for novel actions.
However, our model does not generalize well to in-the-wild objects, subjects, and backgrounds. This is expected, given its training on a modest scale of data. Considering the intricate nature of the generalization to new environments, we regard it as a challenging future research direction.
Nonetheless, 
our focus on generalizing to new actions within the same environment is meaningful for several applications. Some examples: Real-time monitoring during physical rehabilitation (exo-to-ego view translation can help therapists guide patients during rehabilitation sessions), assistive cooking aid for visually impaired individuals (ego views are provided to the visually impaired individuals to cook independently), or real-time fitness training monitoring (coaches use synthesized ego views to help the trainee correct the fitness movements) — in each case, the environment is stable between training and testing, and yet the model must generalize to new actions for cross-view translation.

\paragraph{Object 3D prior} Our empirical qualitative results reveal the limitation of the existing baselines and our Exo2Ego framework when it comes to generating 3D consistent views for novel objects during test time. This can be attributed to the lack of geometric priors for common objects in our daily lives. In order to address this issue, our future works include incorporating robust object geometric priors into our framework, enhancing the realism and precision of the generated objects in ego views, and extending our approach to in-the-wild scenarios.

\paragraph{Action Semantics}
There is currently no dataset available that encompasses pairs of exo-ego videos showcasing a wide spectrum of action semantics. Given the absence of such a dataset, our emphasis in this work lies on the domain of hand-object tabletop interactions. However, our task setting still holds significant value, particularly considering its relevance to a range of applications in augmented reality and robotics.

\end{document}